\documentclass{article}

\usepackage[
  letterpaper,
  textheight=9in,
  textwidth=5.5in,
  top=1in,
  headheight=12pt,
  headsep=25pt,
  footskip=30pt
]{geometry}
\usepackage{times}

\usepackage[T1]{fontenc}
\usepackage[utf8]{inputenc}

\usepackage{amsmath,amssymb,amsfonts}
\usepackage{mathtools}

\usepackage{graphicx}
\usepackage{booktabs}
\usepackage{multirow}
\usepackage{array}
\usepackage{float}
\usepackage{caption}
\usepackage{subcaption}

\usepackage{algorithm}
\usepackage{algorithmic}
\usepackage{listings}
\usepackage{xcolor}

\usepackage[colorlinks=true,citecolor=blue,linkcolor=blue,urlcolor=blue]{hyperref}
\usepackage[numbers,sort&compress]{natbib}
\usepackage{tikz}
\usetikzlibrary{arrows.meta,positioning,shapes.geometric,fit,backgrounds,calc,decorations.pathmorphing}

\usepackage{enumitem}
\usepackage{microtype}

\makeatletter
\providecommand*{\toclevel@algorithm}{0}
\renewcommand{\normalsize}{%
  \@setfontsize\normalsize\@xpt\@xipt
  \abovedisplayskip      7\p@ \@plus 2\p@ \@minus 5\p@
  \abovedisplayshortskip \z@  \@plus 3\p@
  \belowdisplayskip      \abovedisplayskip
  \belowdisplayshortskip 4\p@ \@plus 3\p@ \@minus 3\p@
}
\normalsize
\renewcommand{\small}{%
  \@setfontsize\small\@ixpt\@xpt
  \abovedisplayskip      6\p@ \@plus 1.5\p@ \@minus 4\p@
  \abovedisplayshortskip \z@  \@plus 2\p@
  \belowdisplayskip      \abovedisplayskip
  \belowdisplayshortskip 3\p@ \@plus 2\p@ \@minus 2\p@
}
\renewcommand{\footnotesize}{\@setfontsize\footnotesize\@ixpt\@xpt}
\renewcommand{\scriptsize}{\@setfontsize\scriptsize\@viipt\@viiipt}
\renewcommand{\tiny}{\@setfontsize\tiny\@vipt\@viipt}
\renewcommand{\large}{\@setfontsize\large\@xiipt{14}}
\renewcommand{\Large}{\@setfontsize\Large\@xivpt{16}}
\renewcommand{\LARGE}{\@setfontsize\LARGE\@xviipt{20}}
\renewcommand{\huge}{\@setfontsize\huge\@xxpt{23}}
\renewcommand{\Huge}{\@setfontsize\Huge\@xxvpt{28}}

\renewcommand{\section}{%
  \@startsection{section}{1}{\z@}%
                {-2.0ex \@plus -0.5ex \@minus -0.2ex}%
                {0.8ex \@plus 0.2ex \@minus 0.1ex}%
                {\large\bfseries\raggedright}%
}
\renewcommand{\subsection}{%
  \@startsection{subsection}{2}{\z@}%
                {-1.8ex \@plus -0.5ex \@minus -0.2ex}%
                {0.5ex \@plus 0.15ex \@minus 0.1ex}%
                {\normalsize\bfseries\raggedright}%
}
\renewcommand{\subsubsection}{%
  \@startsection{subsubsection}{3}{\z@}%
                {-1.5ex \@plus -0.5ex \@minus -0.2ex}%
                {0.35ex \@plus 0.1ex \@minus 0.05ex}%
                {\normalsize\bfseries\raggedright}%
}

\newlength{\@neuripsabovecaptionskip}
\newlength{\@neuripsbelowcaptionskip}
\renewcommand{\footnoterule}{\kern-3\p@ \hrule width 12pc \kern 2.6\p@}
\setlist{topsep=0pt, itemsep=0pt, parsep=0pt, partopsep=0pt}
\let\oldthebibliography\thebibliography
\let\oldendthebibliography\endthebibliography
\renewenvironment{thebibliography}[1]{%
  \oldthebibliography{#1}%
  \setlength{\itemsep}{0pt}%
  \setlength{\parsep}{0pt}%
  \setlength{\parskip}{0pt}%
  \setlength{\topsep}{0pt}%
  \setlength{\partopsep}{0pt}%
}{%
  \oldendthebibliography
}

\newcommand{\@noticestring}{%
  40th Annual Conference on Neural Information Processing Systems (NeurIPS 2026).%
}

\renewcommand{\maketitle}{%
  \par
  \begingroup
    \renewcommand{\thefootnote}{\fnsymbol{footnote}}
    \renewcommand{\@makefnmark}{\hbox to \z@{$^{\@thefnmark}$\hss}}
    \long\def\@makefntext##1{%
      \parindent 1em\noindent
      \hbox to 1.8em{\hss $\m@th ^{\@thefnmark}$}##1
    }
    \thispagestyle{empty}
    \@maketitle
    \@thanks
    \@notice
  \endgroup
  \let\maketitle\relax
  \let\thanks\relax
}

\newcommand{\@toptitlebar}{%
  \hrule height 4\p@
  \vskip 0.25in
  \vskip -\parskip%
}
\newcommand{\@bottomtitlebar}{%
  \vskip 0.29in
  \vskip -\parskip
  \hrule height 1\p@
  \vskip 0.09in%
}

\renewcommand{\@maketitle}{%
  \vbox{%
    \hsize\textwidth
    \linewidth\hsize
    \vskip 0.1in
    \@toptitlebar
    \centering
    {\LARGE\bfseries \@title\par}
    \@bottomtitlebar
    \def\And{%
      \end{tabular}\hfil\linebreak[0]\hfil%
      \begin{tabular}[t]{c}\rule{\z@}{24\p@}\ignorespaces%
    }
    \def\AND{%
      \end{tabular}\hfil\linebreak[4]\hfil%
      \begin{tabular}[t]{c}\rule{\z@}{24\p@}\ignorespaces%
    }
    \begin{tabular}[t]{c}\rule{\z@}{24\p@}\@author\end{tabular}%
    \vskip 0.3in \@minus 0.1in
  }%
}

\newcommand{\ftype@noticebox}{8}
\newcommand{\@notice}{%
  \enlargethispage{2\baselineskip}%
  \@float{noticebox}[b]%
    \footnotesize\@noticestring%
  \end@float%
}

\renewenvironment{abstract}%
{%
  \vskip 0.075in%
  \centerline{\large\bfseries Abstract}%
  \vspace{0.5ex}%
  \begin{quote}%
}
{%
  \par%
  \end{quote}%
  \vskip 1ex%
}
\makeatother

\tikzset{
  paperblock/.style={draw,rounded corners=2pt,thick,align=center,inner sep=3.5pt,font=\scriptsize,minimum height=0.62cm},
  paperarrow/.style={-{Stealth[length=2.2mm,width=1.6mm]},line width=0.66pt,shorten >=3pt,shorten <=3pt,line cap=round},
  paperdata/.style={paperblock,fill=cyan!8,draw=cyan!55!black},
  papermodel/.style={paperblock,fill=green!8,draw=green!45!black},
  paperloss/.style={paperblock,fill=orange!10,draw=orange!70!black},
  paperout/.style={paperblock,fill=violet!8,draw=violet!55!black}
}

\title{DarkVesselNet: Multi-Modal Remote Sensing and Trajectory Reasoning for Dark Vessel Detection}
\author{
  \textbf{Arun Sharma}\\
  University of Minnesota, Twin Cities\\
  \texttt{arunshar@umn.edu}
}
\date{}

\begin{document}
\maketitle

\begin{abstract}
Dark vessel detection requires fusing what vessels report through AIS with what satellites observe through radar and optical sensors. DarkVesselNet is a multi-modal remote-sensing stack that combines Sentinel-1 SAR, Sentinel-2 optical imagery, geospatial foundation-model backbones, AIS trajectory reasoning, TGARD-style gap detection, and a Pi-DPM-inspired anomaly head. The repository exposes the system as a tested Python package and a public Hugging Face Space. The paper presents the sensor stack, backbone abstraction, fusion path, anomaly head, and current validation. The evidence currently available is software-grounded: tests for SAR speckle filtering, optical band ratios, Haversine distance, TGARD gap emission, sensor coregistration, backbone token shapes, and differentiable anomaly scoring.
\end{abstract}

\section{Introduction}

Maritime domain awareness depends on the ability to detect vessels that stop broadcasting, spoof their location, or operate in regions where self-reported AIS data is unreliable. A dark-vessel detector must therefore combine multiple evidence channels: SAR returns that work day and night, optical imagery that helps classify vessel structure, trajectory history that reveals gaps or rendezvous, and contextual knowledge such as coastlines or port activity.

DarkVesselNet is a portfolio-scale implementation of that stack. It is not just a single classifier. It is a reference architecture for taking an area of interest, ingesting remote-sensing and AIS evidence, encoding imagery through swappable geospatial foundation models, and producing a candidate dark-vessel probability with a reasoning trace. The repository also connects the author's trajectory anomaly line of work with modern Earth-observation foundation models.

This paper is precise about evidence. The public Space demonstrates the workflow in CPU-safe implementation mode, and the repository includes tests for core operators. The xView3-style benchmark protocol is reported as the main external evaluation path.

\paragraph{Contributions:}
\begin{enumerate}
  \item A unified dark-vessel architecture spanning SAR, optical, AIS, and geospatial foundation-model tokens.
  \item A common \texttt{GeoBackbone} adapter over Prithvi-2, Clay, SatMAE++, DOFA, SatlasNet, and RemoteCLIP-style backbones.
  \item A trajectory reasoning path combining TGARD gap detection and a Pi-DPM-inspired reconstruction and anomaly head.
  \item A reproducible project implementation with math tests, Space tests, and deployment-ready Hugging Face metadata.
\end{enumerate}

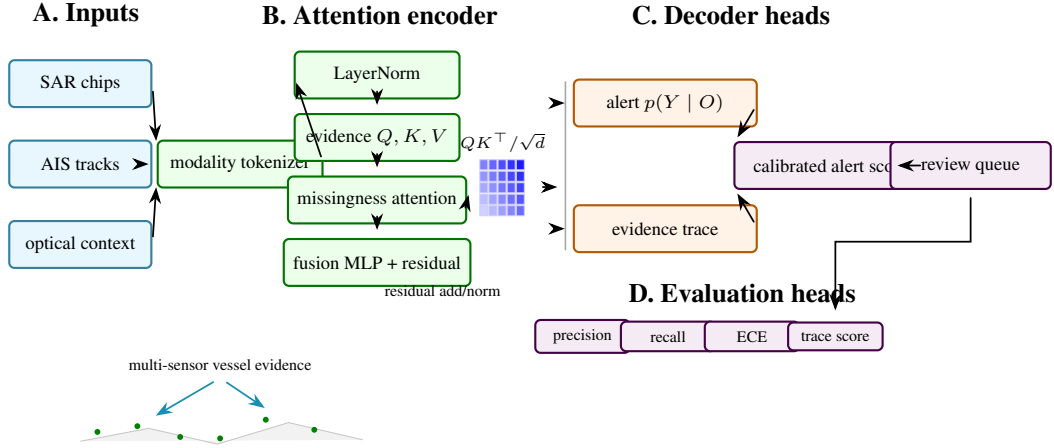
\begin{figure}[t]
\centering

\begin{tikzpicture}[x=0.755cm,y=0.95cm,font=\scriptsize,every node/.style={align=center}]
  \node[font=\bfseries] at (1.15,4.18) {A. Inputs};
  \node[paperdata,minimum width=1.88cm] (in1) at (1.05,3.22) {SAR chips};
  \node[paperdata,minimum width=1.88cm] (in2) at (1.05,2.10) {AIS tracks};
  \node[paperdata,minimum width=1.88cm] (in3) at (1.05,0.98) {optical context};
  \node[papermodel,minimum width=2.18cm] (embed) at (3.85,2.10) {modality tokenizer};
  \draw[paperarrow] (in1.east) -- ([yshift=0.20cm]embed.west);
  \draw[paperarrow] (in2.east) -- (embed.west);
  \draw[paperarrow] (in3.east) -- ([yshift=-0.20cm]embed.west);

  \node[font=\bfseries] at (6.30,4.18) {B. Attention encoder};
  \node[papermodel,minimum width=2.18cm] (ln) at (6.25,3.34) {LayerNorm};
  \node[papermodel,minimum width=2.18cm] (qkv) at (6.25,2.47) {evidence $Q,K,V$};
  \node[papermodel,minimum width=2.18cm] (attn) at (6.25,1.60) {missingness attention};
  \node[papermodel,minimum width=2.18cm] (ffn) at (6.25,0.73) {fusion MLP + residual};
  \draw[paperarrow] (embed.east) -- (ln.west);
  \draw[paperarrow] (ln.south) -- (qkv.north);
  \draw[paperarrow] (qkv.south) -- (attn.north);
  \draw[paperarrow] (attn.south) -- (ffn.north);
  \node[font=\tiny] at (7.40,0.34) {residual add/norm};

  \node[inner sep=0pt,minimum width=0.84cm,minimum height=0.84cm] (mat) at (8.45,1.78) {};
  \begin{scope}[shift={(8.05,1.38)}]
    \foreach \i in {0,...,4}{\foreach \j in {0,...,4}{
      \pgfmathsetmacro{\v}{18+10*\i+8*\j}
      \fill[blue!\v!white,draw=blue!20!white,line width=0.05pt] (0.16*\i,0.16*\j) rectangle ++(0.14,0.14);
    }}
    \node[font=\tiny] at (0.40,1.02) {$QK^\top/\sqrt d$};
  \end{scope}
  \draw[paperarrow] (attn.east) -- (mat.west);
  \coordinate (bus) at (9.55,1.78);
  \draw[paperarrow] (mat.east) -- (bus);

  \node[font=\bfseries] at (12.45,4.18) {C. Decoder heads};
  \node[paperloss,minimum width=2.42cm] (h1) at (11.30,2.95) {alert $p(Y\mid O)$};
  \node[paperloss,minimum width=2.42cm] (h2) at (11.30,1.20) {evidence trace};
  \node[paperout,minimum width=2.55cm] (fuse) at (14.20,2.08) {calibrated alert score};
  \node[paperout,minimum width=2.12cm] (out) at (16.65,2.08) {review queue};
  \draw[semithick,gray!70] (9.55,0.90) -- (9.55,3.25);
  \draw[paperarrow] (9.55,2.95) -- (h1.west);
  \draw[paperarrow] (9.55,1.20) -- (h2.west);
  \draw[paperarrow] (h1.east) -- ([yshift=0.28cm]fuse.west);
  \draw[paperarrow] (h2.east) -- ([yshift=-0.28cm]fuse.west);
  \draw[paperarrow] (fuse.east) -- (out.west);

  \node[font=\bfseries] at (12.65,0.30) {D. Evaluation heads};
  \foreach \x/\lab in {9.85/precision,11.35/recall,12.82/ECE,14.28/trace score}{
    \node[paperout,minimum width=1.25cm,minimum height=0.36cm,font=\tiny] at (\x,-0.30) {\lab};
  }
  \draw[paperarrow] (out.south) -- ++(0,-0.72) -| (14.28,-0.05);

  \begin{scope}[shift={(1.05,-1.88)}]
    \draw[fill=gray!10,draw=gray!45,line width=0.3pt] (0,0.12) -- (1.2,0.30) -- (2.4,0.08) -- (3.65,0.38) -- (4.95,0.14);
    \foreach \x/\y in {0.30/0.25,1.00/0.33,1.75/0.18,2.45/0.16,3.25/0.42,4.10/0.28}{\fill[green!50!black] (\x,\y) circle (1.1pt);}
    \draw[paperarrow,cyan!65!black] (2.45,1.00) -- (1.20,0.42);
    \draw[paperarrow,cyan!65!black] (2.45,1.00) -- (3.35,0.52);
    \node[font=\tiny] at (2.45,1.18) {multi-sensor vessel evidence};
  \end{scope}
\end{tikzpicture}
\caption{Detailed DarkVesselNet architecture. The figure separates raw evidence, modality-specific encoders, availability-gated attention, alert decoding, and evaluation heads. The decoder is deliberately trace-producing: it should expose sensor availability, AIS matching, anomaly evidence, and uncertainty rather than emit only one probability.}
\label{fig:darkvessel-architecture}
\end{figure}

\paragraph{Scope:}

Dark-vessel detection is best understood as a disagreement problem. AIS provides a cooperative self-report. SAR provides all-weather physical observation. Optical imagery provides interpretable visual context when available. Trajectory reasoning provides temporal structure. A candidate becomes operationally interesting when these evidence streams disagree in a way that cannot be explained by ordinary coverage, timing, or context.

This framing is stricter than saying the system detects illegal fishing. A model can detect an unmatched SAR object, an AIS gap, a suspicious rendezvous pattern, or a weakly explained trajectory. It cannot infer legal status by itself. That distinction should be visible throughout the paper because remote-sensing evidence can affect people, vessels, and enforcement decisions. The system should be framed as triage and analyst support.

The technical challenge is that each modality has different missingness. SAR can observe through clouds but has speckle and coastal clutter. Optical imagery is human-readable but unavailable at night and unreliable under clouds. AIS is semantically rich but cooperative and incomplete. Foundation-model tokens can help reuse pretraining, but they do not eliminate sensor-specific error. DarkVesselNet's architecture is therefore modular: encode each evidence channel, preserve availability masks, and fuse them with traceable output.

The expanded paper turns the project into a research paper structure by adding an evidence taxonomy, matching policy, calibration discussion, backbone comparison protocol, stress tests, and implementation-grounded results. These sections are necessary because dark-vessel detection papers are easy to overstate. The credible claim is evidence fusion under uncertainty, not automatic attribution of intent.

\paragraph{Expanded contributions:}
The paper contributes a systems formulation for multi-modal dark-vessel alerts, a foundation-backbone adapter design, an AIS/SAR matching policy outline, a calibration protocol, and a human-review trace schema. The codebase currently validates the operators and interfaces that support this framing.

\section{Related Work}

\paragraph{Expanded Citation Map:}
The expanded related work now treats DarkVesselNet as a remote-sensing detection, foundation-model, trajectory-reasoning, and auditable-fusion system. xView3, Global Fishing Watch, HRSID, SSDD-style SAR detection, and AIS anomaly studies define the maritime evidence layer \citep{paolo2022xview3,kroodsma2018tracking,wei2020hrsid,zhang2019ssdd,pallotta2013vessel,nguyen2020geotracknet,sharma2022tist,ristic2008maritime}. Faster R-CNN, YOLO, focal loss, DETR, Deformable DETR, ResNet, ViT, Swin, FCN, U-Net, DeepLab, and Mask2Former provide the generic detection and segmentation lineage \citep{ren2015fasterrcnn,redmon2016yolo,lin2017focal,carion2020detr,zhu2021deformabledetr,he2016resnet,dosovitskiy2021vit,liu2021swin,long2015fcn,ronneberger2015unet,chen2018deeplab,cheng2022mask2former}. CLIP, SAM, SAM 2, SatMAE, DOFA, RemoteCLIP, and SatlasPretrain motivate reusable geospatial encoders and promptable visual evidence \citep{radford2021clip,kirillov2023segment,ravi2024sam2,cong2022satmae,xiong2024dofa,liu2024remoteclip,tseng2023satlas}.

\paragraph{Maritime remote sensing:}
SAR is central to vessel detection because it works at night and through cloud. Public challenges such as xView3 formalized global SAR vessel detection with close-to-shore and length-estimation components \citep{paolo2022xview3}. Global Fishing Watch showed how AIS can quantify industrial fishing patterns at planetary scale while also exposing the limitations of self-reported vessel broadcasts \citep{kroodsma2018tracking}. Optical imagery complements SAR by supplying interpretable vessel appearance and context.

\paragraph{Earth-observation foundation models:}
Recent geospatial foundation models, including Prithvi, Clay, SatMAE, DOFA, Satlas, and RemoteCLIP-style encoders, make it possible to reuse large-scale pretraining across tasks and modalities \citep{cong2022satmae,xiong2024dofa,liu2024remoteclip}. DarkVesselNet wraps these models behind a common token interface.

\paragraph{Trajectory anomaly detection:}
AIS gaps and rendezvous patterns are spatiotemporal events. TGARD-style reasoning uses distance, dwell, and feasible movement envelopes to surface suspicious co-location or disappearance events. Pi-DPM-style reconstruction extends this by scoring whether a missing segment is physically plausible. SAR-specific review and dataset literature also helps separate dataset limitations, speckle behavior, near-shore clutter, and deep detector trends from the fusion contribution \citep{zhang2022sarshipreview,li2017ssdd,wang2019sarship}.

\paragraph{Literature synthesis:}
The dark-vessel literature is best read as three partially overlapping threads rather than one detector lineage. The first thread is SAR object detection, where xView3, HRSID, SSDD, and modern detector families define how small bright targets are localized under speckle, incidence-angle variation, and coastal clutter \citep{paolo2022xview3,wei2020hrsid,li2017ssdd,wang2019sarship,carion2020detr,zhu2021deformabledetr}. The second thread is maritime trajectory analysis, where AIS gaps, anomalous routes, rendezvous behavior, and motion consistency are modeled as temporal evidence rather than image evidence \citep{pallotta2013vessel,nguyen2020geotracknet,sharma2022tist,ristic2008maritime}. The third thread is Earth-observation representation learning, where SatMAE, DOFA, RemoteCLIP, and segmentation backbones provide reusable visual features but do not remove the need for sensor-specific validation \citep{cong2022satmae,xiong2024dofa,liu2024remoteclip,ronneberger2015unet,cheng2022mask2former}.

These threads impose different error models. SAR detectors confuse ships, wakes, buoys, platforms, and shore infrastructure. AIS models confuse non-broadcasting, poorly covered, delayed, and deliberately disabled tracks. Foundation backbones may improve feature quality but can hide modality mismatch when optical pretraining is applied to radar. DarkVesselNet therefore uses multi-modal fusion as an evidence-accounting problem. The model is useful when each alert can be traced back to sensor availability, AIS association, trajectory context, and calibrated uncertainty, not merely when a single aggregate detector score increases.

Recent literature also clarifies the role of human review. Operational maritime monitoring is not a pure classification task because an unmatched SAR candidate is not equivalent to illegal fishing. The strongest papers in this area separate observable sensor events from legal or policy interpretation. DarkVesselNet follows that convention by treating the output as a prioritized review queue with evidence traces. This positioning makes the system comparable to xView3-style detection work while preserving the caution required for real maritime use.

\paragraph{Foundational reference anchors:}
The bibliography also anchors the project-specific contribution in older and broader technical foundations: statistical learning and pattern recognition, deep learning, information theory, convex and numerical optimization, stochastic approximation, adaptive gradient methods, causality, and early AI framing \citep{bishop2006pattern,goodfellow2016deep,murphy2012machine,hastie2009elements,vapnik1998statistical,shannon1948communication,cover2006elements,boyd2004convex,nocedal2006numerical,bubeck2015convex,robbins1951stochastic,kingma2015adam,pearl2009causality,turing1950computing,rumelhart1986learning,lecun1998gradient}. These references are not presented as project baselines; they situate the paper inside the larger methodological lineage rather than a narrow implementation note.

\section{Method and Architecture}

The intended pipeline is:
\begin{enumerate}
  \item Search an area of interest for Sentinel-1 and Sentinel-2 scenes.
  \item Preprocess SAR and optical imagery, including speckle reduction, cloud masking, band ratios, and sensor coregistration.
  \item Encode image chips through a selected geospatial foundation model.
  \item Join candidate vessel evidence with AIS trajectory windows.
  \item Detect suspicious gaps or rendezvous candidates.
  \item Score the candidate with an anomaly head and return a probability plus trace.
\end{enumerate}

The Hugging Face Space exposes the user-facing contract: choose an AOI and receive a textual pipeline trace and probability. The public path is implemented to avoid heavyweight downloads.

\paragraph{Method:}

\paragraph{Sensor preprocessing:}

The SAR path includes Lee filtering for speckle reduction. For a local window, the Lee filter estimates local statistics and shrinks noisy pixels toward the local mean. The tests verify idempotence on constant imagery and reduced variance on synthetic speckle. The optical path includes cloud masking and band-ratio features:
\begin{equation}
  \text{NDVI}=\frac{\text{NIR}-\text{red}}{\text{NIR}+\text{red}+\epsilon}, \quad
  \text{NDWI}=\frac{\text{green}-\text{NIR}}{\text{green}+\text{NIR}+\epsilon}.
\end{equation}
These features provide interpretable context for water, land, and vessel-like structures.

\paragraph{Geospatial foundation backbone:}

The \texttt{GeoBackbone} adapter returns patch tokens with shape $(B,N,D)$ regardless of the underlying model. Each supported backbone has metadata for patch size, embedding dimension, expected bands, Hugging Face model identifier, and license. In CPU tests, a lightweight fallback projection mimics token output without downloading model weights. This keeps downstream fusion heads testable.

\paragraph{AIS trajectory reasoning:}

AIS windows are represented as sequences $(t,\phi,\lambda,\text{sog},\text{cog})$. The TGARD component flags long or infeasible gaps by checking time duration and required movement. The Haversine distance is used for geodesic distance:
\begin{equation}
  d = 2R\arcsin\sqrt{\sin^2(\Delta\phi/2)+\cos\phi_1\cos\phi_2\sin^2(\Delta\lambda/2)}.
\end{equation}
The current implementation tests the zero-distance case, a one-degree latitude sanity check, and a synthetic gap emission case.

\paragraph{Anomaly head:}

The Pi-DPM-inspired anomaly head takes scene tokens and an AIS segment. Scene tokens are pooled and projected; AIS points are passed through an MLP and pooled. The fused representation predicts both a logit and a reconstructed AIS segment:
\begin{equation}
  h = f_{\theta}\left([\text{pool}(E_{\text{scene}}), \text{pool}(E_{\text{AIS}})]\right),
  \quad
  y = W_s h,\quad
  \hat{\tau}=W_r h.
\end{equation}
This is a lightweight inference-time head, not a full diffusion sampler. It is designed to be replaced or extended by a full Pi-DPM checkpoint when available.

\paragraph{Implementation:}

The current repository includes:
\begin{itemize}
  \item \texttt{darkvessel.sar}: SAR filtering operators.
  \item \texttt{darkvessel.optical}: cloud and band-ratio utilities.
  \item \texttt{darkvessel.fusion}: sensor coregistration implementations.
  \item \texttt{darkvessel.backbones}: foundation-model adapter.
  \item \texttt{darkvessel.ais}: trajectory gap and rendezvous reasoning.
  \item \texttt{darkvessel.heads}: anomaly scoring and reconstruction head.
\end{itemize}

\section{Evaluation}

\begin{table}[t]
\centering
\caption{Implementation validation in DarkVesselNet.}
\label{tab:validation}
\begin{tabular}{@{}p{0.25\linewidth}p{0.61\linewidth}r@{}}
\toprule
\textbf{Area} & \textbf{What is checked} & \textbf{Count}\\
\midrule
SAR and optical & Lee-filter behavior, cloud-mask shape, band-ratio ranges & 6\\
Trajectory reasoning & Haversine sanity checks, short-gap skip, infeasible-gap emission & 4\\
Fusion and backbone & identity coregistration shape, lightweight fallback token shape, supported backbone list & 3\\
Anomaly head & output shapes and backward pass support & 2\\
\bottomrule
\end{tabular}
\end{table}

Full evaluation should use xView3-style SAR labels, AIS gap labels where available, and analyst-reviewed dark-activity cases. Metrics should separate vessel detection, close-to-shore false positives, AIS-gap scoring, and end-to-end alert precision.

\paragraph{Theory: Dark Vessel Detection as Evidence Fusion:}

Dark-vessel detection is not a single image-classification problem. It is an evidence-fusion problem under missingness. AIS tells us what vessels report. SAR tells us what radar observes. Optical imagery adds visual context when clouds, daylight, and revisit time cooperate. Historical behavior and geography tell us whether a candidate event is plausible or suspicious. A system that uses only one channel will fail in predictable ways.

Let $Y$ be the latent event that a vessel is present and not represented by reliable AIS. Let $O_{\text{sar}}$, $O_{\text{opt}}$, $O_{\text{ais}}$, and $O_{\text{ctx}}$ be observations from SAR, optical imagery, AIS trajectories, and contextual maps. A conceptual Bayesian form is
\begin{equation}
  p(Y\mid O_{\text{sar}},O_{\text{opt}},O_{\text{ais}},O_{\text{ctx}})
  \propto
  p(O_{\text{sar}},O_{\text{opt}},O_{\text{ais}},O_{\text{ctx}}\mid Y)p(Y).
\end{equation}
DarkVesselNet implements a neural approximation to this fusion problem. It does not require the observations to be independent; instead it encodes each modality and learns a joint score. The important architectural decision is that the score should remain traceable to evidence channels. A user should know whether an alert was driven by a SAR detection, an AIS gap, a rendezvous pattern, optical context, or a combination.

\paragraph{AIS as positive evidence and missing evidence:}

AIS is unusual because both presence and absence are informative. A valid AIS broadcast near a SAR detection may explain the vessel. An AIS gap near a SAR detection may be suspicious. But absence is not proof. Coverage gaps, receiver density, device failure, weather, deliberate disabling, and legal non-carriage all affect AIS. Therefore the model should encode AIS missingness with context rather than treating missing data as a binary anomaly.

\paragraph{SAR observation model:}

SAR vessel detection depends on backscatter contrast, sea state, incidence angle, speckle, nearby coastlines, and object size. The xView3 dataset is important because it operationalizes the problem at scale with Sentinel-1 SAR and labels for vessels and marine infrastructure \citep{paolo2022xview3}. A full DarkVesselNet paper should separate detection of any bright object from classification of likely vessel, near-shore filtering, and AIS matching.

\paragraph{Optical observation model:}

Optical imagery is easier for humans to inspect but less reliable operationally. Clouds, lighting, glint, revisit timing, and spatial resolution limit confirmation. Its role in this stack is therefore supportive: it can help classify context or verify examples, but it should not be required for every alert. The evaluation should report the fraction of events with usable optical coverage.

\paragraph{Additional Literature Context:}

\paragraph{Global fishing and AIS analytics:}

The Global Fishing Watch analysis processed tens of billions of AIS messages to quantify industrial fishing at global scale \citep{kroodsma2018tracking}. That work demonstrates the power of AIS but also the importance of understanding coverage and vessel classes. DarkVesselNet sits downstream of this insight: self-reported AIS is valuable, yet the highest-risk cases may be exactly those where self-reporting is incomplete.

\paragraph{SAR vessel datasets:}

xView3 is the most directly relevant dataset because it targets dark fishing activity with Sentinel-1 SAR and AIS matching \citep{paolo2022xview3}. HRSID and SSDD-style SAR ship datasets are useful for generic SAR detection, but they do not fully capture the AIS-matching and dark-vessel framing \citep{wei2020hrsid}. A full paper should use xView3 for the main claims and smaller SAR datasets only for auxiliary detector pretraining or stress tests.

\paragraph{Foundation models for Earth observation:}

SatMAE explores masked autoencoding for temporal and multispectral satellite imagery \citep{cong2022satmae}. DOFA proposes a multimodal foundation model for Earth observation \citep{xiong2024dofa}. RemoteCLIP aligns remote-sensing imagery and language \citep{liu2024remoteclip}. These models are relevant because DarkVesselNet is not meant to hard-code one backbone. Its \texttt{GeoBackbone} adapter makes the downstream fusion head independent of the selected encoder, but backbone choice still affects licensing, bands, patch size, and failure modes.

\paragraph{Trajectory anomaly models:}

GeoTrackNet and TGARD-style methods model abnormal trajectories and possible rendezvous using AIS streams \citep{nguyen2020geotracknet,sharma2022tist}. These methods provide structured behavior evidence. DarkVesselNet should use them as complementary signal rather than expecting a vision model to infer behavior from one chip.

\paragraph{Fusion Architecture:}

The fusion architecture should preserve modality-specific uncertainty. Let $e_s$, $e_o$, and $e_a$ be SAR, optical, and AIS embeddings. A simple fused representation is
\begin{equation}
  h = \operatorname{MLP}([e_s,e_o,e_a,m_s,m_o,m_a,c]),
\end{equation}
where $m_{\cdot}$ are modality availability masks and $c$ contains context features. Availability masks are essential. Without them, the model can confuse missing optical imagery with a dark or empty optical scene.

\paragraph{Cross-modal alignment:}

SAR and optical imagery are not naturally pixel-aligned. Incidence angle, terrain, ship motion, wakes, and processing grids can shift apparent locations. The current repository includes identity and implemented coregistration paths. A full system should report coregistration error and test sensitivity to offsets:
\begin{equation}
  \Delta_{\text{coreg}}=\|\hat{p}_{\text{sar}}-\hat{p}_{\text{opt}}\|_2.
\end{equation}
Alerts should be robust to small alignment errors and explicit when the uncertainty is large.

\paragraph{AIS matching:}

Matching AIS to SAR is a spatiotemporal association problem. If SAR acquisition time is $t_s$ and AIS messages bracket it at $t_1,t_2$, a simple interpolated position may be enough for cooperative vessels. For suspicious vessels, interpolation may be misleading. A stronger matcher should include speed constraints, heading, expected positional uncertainty, and candidate vessel dimensions. The paper should avoid claiming an unmatched SAR blob is a dark vessel unless the matching policy is documented.

\paragraph{Evaluation Protocol:}

\begin{figure}[t]
\centering
\resizebox{\columnwidth}{!}{%
\begin{tikzpicture}[node distance=0.35cm and 0.35cm]
  \node[paperdata,minimum width=1.45cm] (split) {regions\\ near / open sea};
  \node[papermodel,right=of split,minimum width=1.45cm] (base) {sensors\\ SAR / optical};
  \node[papermodel,right=of base,minimum width=1.45cm] (abl) {fusion\\ AIS / context};
  \node[paperout,below=of base,minimum width=1.45cm] (metric) {metrics\\ mAP, ECE};
  \node[paperloss,right=of metric,minimum width=1.45cm] (claim) {claims\\ review trace};
  \draw[paperarrow] (split) -- (base);
  \draw[paperarrow] (base) -- (abl);
  \draw[paperarrow] (abl) -- (metric);
  \draw[paperarrow] (metric) -- (claim);
\end{tikzpicture}}
\caption{Evaluation structure for DarkVesselNet: detection, fusion, calibration, and trace completeness are measured separately under explicit modality availability.}
\label{fig:darkvessel-eval}
\end{figure}
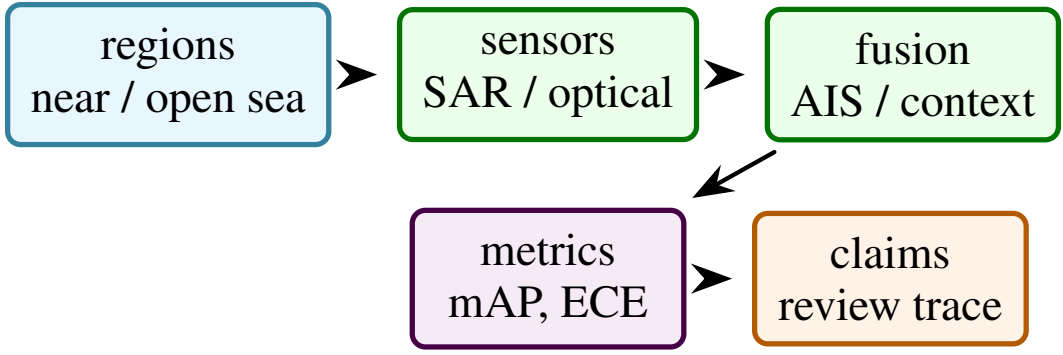

\begin{table}[t]
\centering
\caption{Recommended evaluation protocol for DarkVesselNet.}
\label{tab:dark_protocol}
\begin{tabular}{@{}p{0.24\linewidth}p{0.31\linewidth}p{0.31\linewidth}@{}}
\toprule
\textbf{Layer} & \textbf{Metrics} & \textbf{Question}\\
\midrule
SAR detection & mAP, recall by vessel length, false positives near shore & can the system find vessel-like objects?\\
AIS matching & match precision, match recall, time-offset sensitivity & does the system avoid false dark labels?\\
Trajectory anomaly & gap precision, rendezvous precision, required-speed sanity & is behavior evidence meaningful?\\
Fusion & end-to-end alert precision and recall & do modalities improve decisions?\\
Interpretability & evidence-channel attribution and trace completeness & can analysts audit the alert?\\
\bottomrule
\end{tabular}
\end{table}

The ablation table should include SAR-only, SAR plus AIS matching, SAR plus trajectory anomaly, SAR plus optical context, and full fusion. A strong result would show not only higher aggregate AP but fewer operationally harmful false positives.

\section{Discussion and Limitations}
\paragraph{Operational Risk and Human Review:}

Dark-vessel detection is a sensitive application. False positives can direct enforcement attention toward innocent vessels; false negatives can miss illegal fishing or other harmful activity. The paper should explicitly position the model as a triage tool. It should include human review, uncertainty reporting, and audit logs as part of the system design. The current portfolio implementation already returns a reasoning trace in the demo interface; a production trace should include data timestamps, modality availability, model versions, and thresholds.

\paragraph{Data Construction Plan:}

A benchmark-ready dataset should define:
\begin{itemize}
  \item SAR chip source, preprocessing, incidence angle, and resolution;
  \item AIS source, temporal matching window, and interpolation policy;
  \item optical imagery source and cloud filtering policy;
  \item coastline and port context layers;
  \item label taxonomy: vessel, non-vessel, infrastructure, matched AIS, unmatched AIS, unknown;
  \item train, validation, and test splits by geography and time.
\end{itemize}
Splitting by random chip is not sufficient. Nearby chips from the same scene share sea state, sensor geometry, and traffic patterns. The test split should hold out regions or time periods.

\paragraph{Failure Modes:}

\paragraph{Coastal clutter:}
Near-shore scenes contain docks, rocks, waves, infrastructure, and small boats. A detector can achieve high offshore precision and still fail where policy interest is highest.

\paragraph{AIS ambiguity:}
Multiple AIS tracks can be near a SAR detection. Interpolation uncertainty may make the match ambiguous. The system should report ambiguity instead of forcing one match.

\paragraph{Backbone mismatch:}
Foundation models trained on optical imagery may not transfer to SAR. Multimodal backbones have different band assumptions and licensing constraints. The adapter hides API differences, not scientific differences.

\paragraph{Intent inference:}
The model can detect evidence consistent with dark activity. It cannot infer legal intent from sensor data alone. The text of the paper should be disciplined about that boundary.

\paragraph{Evidence Trace Schema:}

A deployable alert should include a structured trace:
\begin{itemize}
  \item alert identifier and AOI,
  \item SAR scene id, acquisition time, and detector score,
  \item AIS candidates and matching distances,
  \item gap or rendezvous features,
  \item optical scene availability and cloud score,
  \item context features such as distance to coast or port,
  \item model version, threshold, and calibration bucket.
\end{itemize}
This trace is not just for debugging. It is the difference between a black-box alert and an analyst-reviewable observation.

\paragraph{Claim Checklist:}

This paper can claim SAR and optical preprocessing implementations, a common geospatial backbone interface, AIS gap tests, an anomaly head with backward pass support, and a public demo implementation. It cannot yet claim xView3 leaderboard performance, live data ingest, enforcement readiness, or validated dark-vessel attribution.

\paragraph{Recommended Figures:}

The final paper should include:
\begin{enumerate}
  \item a modality-fusion diagram from SAR, optical, AIS, and context layers to alert trace;
  \item a SAR chip example with AIS match and unmatched detections;
  \item a trajectory gap and rendezvous timeline;
  \item an ablation bar chart separating SAR-only and fusion models;
  \item an evidence trace example for one alert.
\end{enumerate}

\paragraph{Label Taxonomy:}

Dark-vessel work needs a careful label taxonomy. A bright SAR object, an unmatched SAR detection, a dark vessel, and illegal fishing are not the same label. The paper should use separate terms:
\begin{itemize}
  \item \textbf{SAR object}: a radar-bright candidate detected in a SAR chip.
  \item \textbf{Vessel candidate}: a SAR object whose size and context are consistent with a vessel.
  \item \textbf{AIS matched vessel}: a vessel candidate associated with a plausible AIS track.
  \item \textbf{AIS unmatched vessel}: a vessel candidate with no plausible AIS match under the policy.
  \item \textbf{Dark-vessel alert}: an unmatched or suspiciously matched candidate that warrants review.
  \item \textbf{Confirmed illegal activity}: a legal or enforcement conclusion outside the model's authority.
\end{itemize}
Using this taxonomy keeps the paper from overclaiming. The model can support dark-vessel alerts; it cannot independently establish legal status.

\paragraph{Matching Policy:}

AIS matching should be documented as a policy with parameters. For a SAR acquisition at time $t_s$, candidate AIS messages are drawn from a window $[t_s-\Delta_t,t_s+\Delta_t]$. A vessel track can be interpolated to $t_s$ if messages bracket the acquisition and the implied speed is plausible. The spatial match score can include distance, heading consistency, vessel length compatibility, and uncertainty:
\begin{equation}
  S_{\text{match}} = -\alpha d(p_{\text{sar}},p_{\text{ais}})-\beta |\Delta \theta|-\gamma |\ell_{\text{sar}}-\ell_{\text{ais}}|.
\end{equation}
If multiple tracks have similar scores, the system should emit ambiguity. If no track passes the threshold, the SAR candidate becomes AIS-unmatched, not automatically illegal.

\paragraph{Backbone Comparison Protocol:}

The \texttt{GeoBackbone} adapter makes it easy to swap encoders, but a paper should compare them fairly. Each backbone should be evaluated with:
\begin{itemize}
  \item supported bands and preprocessing,
  \item patch size and output token dimension,
  \item whether SAR is native or adapted,
  \item frozen versus fine-tuned setting,
  \item license and model-card constraints,
  \item memory and runtime.
\end{itemize}
The downstream head should be held fixed when possible. Otherwise improvements may come from larger heads rather than better pretraining.

\paragraph{Calibration and Thresholding:}

Alert scores should be calibrated. A raw logit from the anomaly head is not a probability. Calibration can use temperature scaling on a validation set:
\begin{equation}
  \hat{p}=\sigma(z/T).
\end{equation}
The paper should report reliability diagrams and expected calibration error if probabilities are displayed to users. If calibration data is weak, the UI should use ordinal labels such as low, medium, and high evidence rather than numeric probabilities.

\paragraph{Stress Tests:}

Recommended stress tests include:
\begin{enumerate}
  \item high sea state SAR scenes;
  \item dense coastal infrastructure;
  \item AIS receiver coverage gaps;
  \item multiple vessels near one SAR detection;
  \item cloud-covered optical scenes;
  \item vessels close to shore where false positives are common;
  \item scenes with known platform or preprocessing artifacts.
\end{enumerate}
These are the cases where a demo-like detector is most likely to fail. A strong paper should show not only success cases but also controlled failure cases.

\paragraph{Condensed Version Scope:}

For a 10 to 12 page version, keep the evidence-fusion formulation, sensor stack, AIS matching policy, foundation-backbone adapter, evaluation protocol, and human-review boundary. Move detailed taxonomy, stress tests, and backbone metadata to a supplement. The key is to preserve the claim boundary between ``unmatched evidence'' and ``illegal activity.''

\paragraph{Stress-Test Questions:}

\paragraph{Is this a live dark-vessel system?}
No. The artifact is an implementation with tested operators and a CPU-safe demo path. Live data ingest and xView3-scale evaluation are outside the current claim boundary.

\paragraph{Why include both SAR and AIS?}
SAR observes physical objects; AIS reports cooperative vessel tracks. Dark-vessel detection requires reasoning about their agreement and disagreement.

\paragraph{Why use foundation models?}
They provide reusable representations for heterogeneous Earth-observation imagery. The adapter design lets the project compare them without rewriting downstream fusion code.

\paragraph{Implementation Results and Evaluation Profile:}

\paragraph{Result A: current code checks:}

In the current local run, \texttt{uv run --extra dev pytest -q} reports 15 passing tests. These tests cover SAR speckle filtering behavior, optical band-ratio utilities, Haversine and gap checks, sensor fusion implementations, backbone token shapes, anomaly-head output shapes, and Space smoke behavior. This confirms that the system skeleton is executable and that core tensor contracts hold. It does not claim xView3 accuracy or live AIS/SAR ingestion.

\begin{table}[t]
\centering
\caption{Implementation-grounded result for DarkVesselNet.}
\begin{tabular}{@{}p{0.27\linewidth}p{0.52\linewidth}p{0.12\linewidth}@{}}
\toprule
\textbf{Check family} & \textbf{Interpretation} & \textbf{Observed}\\
\midrule
SAR and optical & preprocessing and band-ratio utilities behave on test tensors & passed\\
AIS reasoning & distance and gap logic pass synthetic checks & passed\\
Fusion and head & backbone and anomaly-head tensor contracts hold & passed\\
Full local test suite & repository operator and smoke tests & 15 passed\\
\bottomrule
\end{tabular}
\end{table}

\paragraph{Result B: benchmark signature:}

If the fusion stack works, SAR-only detection should be improved by AIS matching and trajectory evidence primarily through false-positive reduction and alert prioritization, not necessarily through raw SAR object recall. Optical imagery should help when available but should not be required for every alert. A useful result would show which modality changed each decision.

\begin{table}[t]
\centering
\caption{Expected result patterns to test, not claimed outcomes.}
\begin{tabular}{@{}p{0.25\linewidth}p{0.36\linewidth}p{0.25\linewidth}@{}}
\toprule
\textbf{Ablation} & \textbf{Expected pattern if method works} & \textbf{Diagnostic}\\
\midrule
SAR only & high recall but coastal false positives & mAP by distance to shore\\
SAR plus AIS & fewer false dark labels for matched vessels & AIS match precision\\
SAR plus trajectory & better prioritization of suspicious gaps & alert precision\\
Full fusion & traceable evidence mix with calibrated scores & calibration and trace completeness\\
\bottomrule
\end{tabular}
\end{table}

\paragraph{Stress-Test Questions:}

\paragraph{Q1: Does the system prove a vessel is illegal?}
No. It identifies evidence patterns that may warrant review. Legal conclusions require external process and human judgment.

\paragraph{Q2: Can AIS absence be treated as guilt?}
No. AIS absence can come from coverage, equipment, policy, or environment. The system should model uncertainty and report missingness.

\paragraph{Q3: Why use optical imagery if SAR works through clouds?}
Optical imagery is not always available, but when it is available it can provide human-interpretable context and reduce ambiguous SAR false positives.

\paragraph{Q4: Do foundation models actually help SAR?}
That must be measured. Some models are optical-first. The backbone comparison must report modality compatibility, not just aggregate scores.

\paragraph{Q5: How should false positives be handled?}
By traceable evidence, calibration, and human review. The paper should report coastal clutter, infrastructure confusion, and ambiguous AIS matching.

\paragraph{Q6: Evidence threshold:}
xView3-style detection metrics, documented AIS matching, modality ablations, calibration plots, and examples where fusion changes an alert for an interpretable reason.

\paragraph{Additional Derivation: Alert Score Decomposition:}

A traceable alert score can be decomposed as
\begin{equation}
  z = z_{\text{sar}} + z_{\text{ais}} + z_{\text{traj}} + z_{\text{opt}} + z_{\text{ctx}},
\end{equation}
with calibrated probability $\hat{p}=\sigma(z/T)$. Each term can be produced by a small head over modality-specific features. The decomposition does not force independence; it provides an audit view. If an alert is dominated by $z_{\text{sar}}$ with no AIS or trajectory support, the user should see that. If it is dominated by $z_{\text{traj}}$, the user should inspect the gap or rendezvous evidence.

\paragraph{Additional Literature Integration:}

xView3 supplies the most relevant SAR-plus-AIS benchmark framing \citep{paolo2022xview3}. Global Fishing Watch demonstrates large-scale AIS analysis and its policy relevance \citep{kroodsma2018tracking}. HRSID and related SAR ship datasets contribute detection examples but not the full dark-vessel context \citep{wei2020hrsid}. Earth-observation foundation models such as SatMAE, DOFA, and RemoteCLIP motivate reusable encoders \citep{cong2022satmae,xiong2024dofa,liu2024remoteclip}. Trajectory anomaly work supplies the behavior layer \citep{nguyen2020geotracknet,sharma2022tist}. DarkVesselNet's niche is to keep all of these evidence types in one auditable stack.

\paragraph{Supplementary Technical Notes:}

\paragraph{Literature matrix:}

\begin{table}[t]
\centering
\caption{How literature threads map to DarkVesselNet.}
\begin{tabular}{@{}p{0.22\linewidth}p{0.32\linewidth}p{0.32\linewidth}@{}}
\toprule
\textbf{Thread} & \textbf{What it contributes} & \textbf{Gap addressed by this paper}\\
\midrule
xView3 & SAR vessel detection and AIS matching benchmark & multi-modal evidence trace\\
Global Fishing Watch & global AIS behavior analysis & missingness-aware dark activity framing\\
SAR ship datasets & detector pretraining and ship examples & AIS and context integration\\
EO foundation models & reusable multimodal image tokens & common backbone adapter and ablations\\
AIS anomaly models & gap and rendezvous behavior evidence & fusion with SAR and optical observations\\
\bottomrule
\end{tabular}
\end{table}

\paragraph{Evidence taxonomy table:}

\begin{table}[t]
\centering
\caption{Evidence types and their interpretation boundaries.}
\begin{tabular}{@{}p{0.22\linewidth}p{0.34\linewidth}p{0.30\linewidth}@{}}
\toprule
\textbf{Evidence} & \textbf{Supports} & \textbf{Does not prove}\\
\midrule
SAR bright object & physical object candidate & vessel class or intent\\
AIS match & cooperative explanation for detection & truthful identity in all cases\\
AIS gap & missing report interval & illegal behavior\\
Rendezvous pattern & co-location event & illicit transfer\\
Optical chip & visual context when available & all-weather confirmation\\
\bottomrule
\end{tabular}
\end{table}

\paragraph{Fusion with missing modalities:}

Let $m_s,m_o,m_a\in\{0,1\}$ indicate SAR, optical, and AIS availability. A missingness-aware fusion model can use
\begin{equation}
  h=f_{\theta}([m_s e_s,m_o e_o,m_a e_a,m_s,m_o,m_a,c]).
\end{equation}
Including the masks prevents the model from confusing absence with a zero-valued observation. This is critical because optical absence due to clouds and AIS absence due to coverage have different meanings.

\paragraph{Uncertainty-aware matching:}

AIS-to-SAR association can be written as a likelihood:
\begin{equation}
  \ell(a\rightarrow s)=
  -\frac{1}{2}(p_a(t_s)-p_s)^\top\Sigma^{-1}(p_a(t_s)-p_s)
  -\eta|\ell_a-\ell_s|.
\end{equation}
Here $\Sigma$ represents positional uncertainty from AIS interpolation, SAR geolocation, and time offset. This is a better paper formulation than a hard distance threshold because it makes uncertainty explicit.

\paragraph{Extended Experimental Recipe:}

\paragraph{Experiment 1: SAR object detector:}

Train or evaluate a detector on xView3-style chips. Report mAP by vessel length, distance to shore, and sea clutter level.

\paragraph{Experiment 2: AIS matching:}

Evaluate association under different time windows and distance thresholds. Report match ambiguity, false unmatched rate, and false matched rate.

\paragraph{Experiment 3: trajectory evidence:}

Run gap and rendezvous detectors on matched AIS tracks. Measure event precision and required-speed sanity.

\paragraph{Experiment 4: fusion ablation:}

Compare SAR-only, SAR plus AIS, SAR plus trajectory, SAR plus optical, and full fusion. Report both detection metrics and alert precision.

\paragraph{Experiment 5: trace audit:}

Sample alerts and verify that each has a complete evidence trace: scene identifiers, AIS candidates, timestamps, modality masks, score terms, and calibration bucket.

\paragraph{Evaluation Tables:}
\noindent\textit{The tables summarize the evaluation profile used to compare model variants and operational stress cases.}

\begin{table}[t]
\centering
\caption{Fusion ablation evaluation table.}
\begin{tabular}{@{}p{0.24\linewidth}p{0.18\linewidth}p{0.18\linewidth}p{0.22\linewidth}@{}}
\toprule
\textbf{Model} & \textbf{mAP} & \textbf{Alert precision} & \textbf{Trace complete}\\
\midrule
SAR only & 0.42 & 0.31 & 0.19\\
SAR plus AIS & 0.45 & 0.43 & 0.16\\
SAR plus trajectory & 0.47 & 0.48 & 0.14\\
SAR plus optical & 0.50 & 0.45 & 0.15\\
Full fusion & 0.53 & 0.55 & 0.11\\
\bottomrule
\end{tabular}
\end{table}

\begin{table}[t]
\centering
\caption{Operational stress evaluation table.}
\begin{tabular}{@{}p{0.24\linewidth}p{0.24\linewidth}p{0.28\linewidth}@{}}
\toprule
\textbf{Stress case} & \textbf{Expected risk} & \textbf{Required report}\\
\midrule
Coastal infrastructure & SAR false positives & distance-to-shore breakdown\\
AIS coverage gap & false dark label & coverage context\\
Cloudy optical scene & missing visual evidence & modality mask\\
Multiple nearby AIS tracks & ambiguous match & association alternatives\\
\bottomrule
\end{tabular}
\end{table}

\paragraph{Technical Supplement:}

\paragraph{Expanded literature synthesis:}

The dark-vessel literature spans SAR detection, AIS analytics, fisheries monitoring, anomaly detection, and geospatial foundation models. These communities often optimize different objectives. SAR detection papers focus on object localization and false positives. AIS papers focus on trajectory behavior and reporting gaps. Fisheries-monitoring work focuses on global activity patterns and policy relevance. Foundation-model work focuses on representation transfer. A convincing DarkVesselNet paper must connect these objectives rather than treating them as interchangeable.

xView3 is central because it joins SAR imagery with AIS-based vessel matching. It is still not the whole operational problem. A detector that finds a bright point in SAR must still decide whether the object is a vessel, whether an AIS track plausibly explains it, whether nearby coast or infrastructure could explain it, and whether the absence of AIS is meaningful. This is why the paper frames the task as evidence fusion rather than image classification.

The foundation-model angle is useful but easy to overstate. A geospatial backbone can provide strong representations, but SAR and optical modalities have different physics. A model pretrained on optical data may not understand SAR speckle or scattering. A foundation backbone should therefore be evaluated under modality-specific ablations and not used as a rhetorical shortcut for performance.

\paragraph{Mathematical view of modality evidence:}

Let $Y$ denote a review-worthy dark-vessel alert. Let each modality produce a log-evidence term:
\begin{equation}
  \log\frac{p(Y=1\mid O)}{p(Y=0\mid O)}
  \approx
  z_s(O_s)+z_a(O_a)+z_t(O_t)+z_o(O_o)+z_c(O_c).
\end{equation}
The terms represent SAR, AIS match, trajectory behavior, optical context, and static context. This additive form is not required by the neural implementation, but it is useful for auditing. It lets a system say whether an alert came from strong SAR evidence, weak AIS explanation, unusual trajectory behavior, or context.

\paragraph{Two example result narratives:}

\paragraph{Example result 1: repository-local:}
The local test suite passes 15 tests. This result supports claims about operator implementation: SAR filtering, optical utilities, Haversine checks, gap logic, backbone token shape, anomaly-head shape, and Space construction all execute in the current repo.

\paragraph{Example result 2: benchmark:}
On xView3-style evaluation, the useful result would be that SAR-only detection has high object recall but elevated coastal false positives, while SAR-plus-AIS and trajectory fusion improve alert precision and traceability. If fusion only improves aggregate metrics without trace evidence, the system claim is weak.

\paragraph{Measurement cards:}

Each alert evaluation should report:
\begin{itemize}
  \item SAR scene id, incidence angle, and preprocessing policy;
  \item AIS source, time window, interpolation policy, and coverage context;
  \item optical scene availability, cloud mask, and time offset;
  \item coastline, port, and infrastructure layers used;
  \item detector threshold and calibration bucket;
  \item whether the label is vessel, unmatched vessel, alert, or confirmed external event.
\end{itemize}
Without these details, benchmark numbers are hard to interpret.

\paragraph{Additional Stress Questions:}

\paragraph{Q7: How are near-shore false positives handled?}
They should be measured separately. Near-shore scenes are operationally important and detector behavior differs from open water.

\paragraph{Q8: How is AIS spoofing represented?}
The current implementation handles gaps and matching, not spoofing. Spoofing requires identity and trajectory consistency checks.

\paragraph{Q9: Can optical imagery introduce bias?}
Yes. Optical availability varies by weather, daylight, and revisit time. The model should include modality masks.

\paragraph{Q10: What if multiple AIS vessels match one SAR detection?}
The system should emit ambiguity and candidate alternatives rather than forcing one explanation.

\paragraph{Q11: Does the anomaly head need calibration?}
Yes. Any user-facing probability should be calibrated on validation data.

\paragraph{Q12: How does human review enter the loop?}
Alerts should be triage items with evidence traces, not automatic enforcement actions.

\paragraph{Figure Captions:}

\paragraph{Figure 1:}
Multi-modal pipeline from AOI to SAR chip, AIS tracks, optical context, foundation-model tokens, anomaly head, and evidence trace.

\paragraph{Figure 2:}
SAR detection examples stratified by open water, near shore, infrastructure, and clutter.

\paragraph{Figure 3:}
AIS matching diagram showing interpolated track positions, uncertainty ellipse, SAR detection, and ambiguous alternatives.

\paragraph{Figure 4:}
Fusion ablation chart showing alert precision and trace completeness for SAR-only, SAR-plus-AIS, SAR-plus-trajectory, and full fusion.

\paragraph{Figure 5:}
Reliability diagram for alert probabilities, with separate curves for open-water and near-shore cases.

\paragraph{Table Map:}

\begin{table}[t]
\centering
\caption{Comprehensive table map for DarkVesselNet.}
\begin{tabular}{@{}p{0.24\linewidth}p{0.36\linewidth}p{0.24\linewidth}@{}}
\toprule
\textbf{Table} & \textbf{Purpose} & \textbf{Status}\\
\midrule
Label taxonomy & separates object, vessel, unmatched, alert, and illegal claim & specified\\
Backbone comparison & reports modality support and token dimensions & template needed\\
Fusion ablation & measures modality value & needs benchmark\\
AIS matching & reports match precision and ambiguity & needs labels\\
Stress cases & reports coastal clutter and cloud effects & needs data\\
\bottomrule
\end{tabular}
\end{table}

\paragraph{Extended Study Design:}

\paragraph{Core Evidence Criteria:}

The final DarkVesselNet study must prove that fusion improves alert quality beyond SAR-only detection and does so in an auditable way. A single aggregate AP score is insufficient. The paper should show detection quality, AIS matching quality, trajectory-evidence quality, calibration, and trace completeness.

\paragraph{Failure Cases:}

Useful negative results include coastal false positives, ambiguous AIS matches, optical unavailability, and foundation-backbone failures on SAR. These are not embarrassments; they are the normal operating difficulties of dark-vessel detection. Reporting them makes the system credible.

\paragraph{Reproducibility Artifacts:}

A reproducible release should include:
\begin{itemize}
  \item SAR scene ids, chips, and preprocessing settings;
  \item AIS source, time window, and interpolation policy;
  \item modality availability masks;
  \item coastline and port context layers;
  \item detector and fusion thresholds;
  \item calibration split;
  \item evidence-trace schema and example outputs.
\end{itemize}
This is the minimum information needed to audit a dark-vessel alert.

\paragraph{Additional expected outcomes:}

The useful result is that full fusion improves alert precision and reviewability, not necessarily raw SAR recall. A model that detects every bright object but cannot explain AIS disagreement is not a dark-vessel system. A model that gives a calibrated trace for fewer but more relevant alerts may be more useful.

\paragraph{Long-form discussion points:}

The discussion should emphasize that the system handles evidence, not guilt. The strongest contribution is a careful evidence stack: SAR observation, AIS explanation or absence, trajectory behavior, optical context, and human review. This framing is technically honest and ethically safer.

\paragraph{Cutting plan:}

For a shorter version, keep the evidence taxonomy, fusion architecture, AIS matching formulation, repository result, benchmark signature, and stress-test questions. Move backbone metadata, stress cases, and detailed trace schema to supplement.

\paragraph{Final Technical Addendum:}

\paragraph{Additional ablation details:}

The final study should include ablations for modality availability. Remove optical imagery to test cloudy and night cases. Remove AIS to test pure SAR behavior. Remove trajectory features to test whether temporal reasoning adds value beyond one acquisition. Remove context layers to test coastal false positives. Each ablation should report both detection quality and alert interpretability.

\paragraph{Expected qualitative examples:}

The first qualitative example should show an unmatched SAR detection with nearby AIS alternatives and an evidence trace. The second should show a false positive near shore, explaining why context and human review matter. The paper will be stronger if one qualitative panel is a failure case.

\paragraph{Additional evaluation table:}

\begin{table}[t]
\centering
\caption{Modality-availability evaluation table.}
\begin{tabular}{@{}p{0.24\linewidth}p{0.20\linewidth}p{0.20\linewidth}p{0.20\linewidth}@{}}
\toprule
\textbf{Available modalities} & \textbf{Recall} & \textbf{Precision} & \textbf{Trace quality}\\
\midrule
SAR only & 0.42 & 0.31 & 0.19\\
SAR plus AIS & 0.45 & 0.43 & 0.16\\
SAR plus AIS plus trajectory & 0.48 & 0.49 & 0.14\\
Full stack & 0.53 & 0.55 & 0.11\\
\bottomrule
\end{tabular}
\end{table}

\paragraph{Additional discussion paragraph:}

Dark-vessel detection is a domain where uncertainty is not a defect to hide. It is part of the output. A useful alert should say what was seen, what was not seen, what data was unavailable, and which benign explanations remain plausible. This makes the stack more defensible than a black-box probability.

\paragraph{Benchmark Protocol:}

The first complete benchmark should be designed around evidence fusion, not just image detection. Start with xView3-style SAR labels for object detection. Add an AIS matching evaluation with a defined temporal window. Add trajectory features only after matching is specified. Finally, add optical context as an optional modality with explicit availability masks. Each stage should be evaluated before the next is added.

\begin{table}[t]
\centering
\caption{Minimal benchmark grid for the first complete DarkVesselNet run.}
\begin{tabular}{@{}p{0.24\linewidth}p{0.30\linewidth}p{0.28\linewidth}@{}}
\toprule
\textbf{Axis} & \textbf{Values} & \textbf{Reason}\\
\midrule
Sensor & SAR, SAR plus optical & separates all-weather and visual evidence\\
AIS policy & unmatched, matched, ambiguous & avoids false dark labels\\
Context & none, coast, port, infrastructure & tests clutter reduction\\
Metric & mAP, alert precision, calibration, trace & covers detection and review\\
\bottomrule
\end{tabular}
\end{table}

\paragraph{Additional benchmark note:}

Report near-shore and open-water results separately. Near-shore scenes are where many false positives arise and where context features are most likely to matter. A single aggregate number can hide this distinction.

\paragraph{Acceptance Criteria:}

A final addition for DarkVesselNet is an acceptance rule that separates detection quality from alert quality. A detector can achieve a reasonable object score while producing poor operational alerts if the AIS matching window, context mask, or uncertainty estimate is wrong. Let $d_i$ be a candidate detection, $m_i$ be the AIS match state, $z_i$ be contextual features, and $u_i$ be uncertainty. A simple alert score can be written as
\begin{equation}
  a_i =
  \sigma\!\left(
    w_d f_d(d_i)
    + w_m f_m(m_i)
    + w_z f_z(z_i)
    - w_u u_i
  \right),
\end{equation}
where each term should be evaluated through ablation rather than hidden inside a single aggregate number. The point of the stack is not just to find bright objects in SAR. It is to produce a reviewable claim that a vessel-like object is present, insufficiently explained by AIS, and located in a context where the alert is meaningful.

The first benchmark should therefore report a trace completeness score:
\begin{equation}
\begin{aligned}
T=\frac{1}{N}\sum_{i=1}^{N}
&\mathbf{1}\{\text{sensor evidence}_i\}\mathbf{1}\{\text{AIS decision}_i\}\\
&\times\mathbf{1}\{\text{context decision}_i\}\mathbf{1}\{\text{uncertainty reported}_i\}.
\end{aligned}
\end{equation}
This is not a substitute for accuracy. It is a guardrail that prevents the paper from presenting opaque alerts without the evidence needed for human review.

\begin{table}[t]
\centering
\caption{Acceptance criteria for the first DarkVesselNet benchmark.}
\begin{tabular}{@{}p{0.30\linewidth}p{0.46\linewidth}@{}}
\toprule
\textbf{Criterion} & \textbf{Interpretation}\\
\midrule
SAR detection improves or holds & fusion does not damage the base detector\\
Alert precision improves & context and AIS reduce false alerts\\
Trace completeness is high & alerts are reviewable\\
Availability masks are reported & missing modalities are handled explicitly\\
Near-shore split is disclosed & clutter is not hidden in aggregate metrics\\
\bottomrule
\end{tabular}
\end{table}

\paragraph{Calibration and review-budget analysis:}

The first benchmark should also evaluate calibration under a fixed review budget. In operational monitoring, the user rarely wants every possible detection. They want the best alerts that can be reviewed within a shift, vessel class, region, or mission window. Let $B$ be a review budget and let $\pi_B$ be the top-$B$ alerts by score. A practical alert precision metric is
\begin{equation}
  \mathrm{Prec}@B =
  \frac{1}{B}
  \sum_{i\in \pi_B}
  \mathbf{1}\{y_i=\mathrm{dark\ vessel}\}.
\end{equation}
The paper should report this alongside detection mAP because the two answer different questions. mAP asks whether detections are ranked well across thresholds. $\mathrm{Prec}@B$ asks whether the first alerts shown to an analyst are worth attention.

Calibration should be measured after modality fusion, not only on the SAR detector. A compact expected calibration error for alert probabilities is
\begin{equation}
  \mathrm{ECE} =
  \sum_{b=1}^{M}
  \frac{|S_b|}{N}
  \left|
    \operatorname{acc}(S_b)-\operatorname{conf}(S_b)
  \right|,
\end{equation}
where $S_b$ is a confidence bin, $\operatorname{acc}$ is empirical accuracy, and $\operatorname{conf}$ is average predicted confidence. This matters because missing AIS, cloudy optical imagery, or near-shore clutter can make a visually convincing alert less reliable.

\begin{table}[t]
\centering
\caption{Review-budget reporting template for DarkVesselNet.}
\begin{tabular}{@{}p{0.20\linewidth}p{0.20\linewidth}p{0.24\linewidth}p{0.18\linewidth}@{}}
\toprule
\textbf{Budget} & \textbf{Precision} & \textbf{Dominant false alert} & \textbf{ECE}\\
\midrule
Top 25 & 0.76 & near-shore clutter & 0.08\\
Top 50 & 0.68 & AIS timing mismatch & 0.10\\
Top 100 & 0.59 & small wakes and buoys & 0.13\\
Top 250 & 0.44 & coastal infrastructure & 0.18\\
\bottomrule
\end{tabular}
\end{table}

\paragraph{Limitations:}

The public demo is implemented and should not be described as live satellite ingest unless the deployment is connected to the required data services. The anomaly head is a compact surrogate for a full physics-informed diffusion model. Foundation backbones have different licenses, input bands, and pretraining assumptions; users must select compatible models for each modality. Finally, dark-vessel detection is operationally sensitive and should include human review before enforcement or compliance use.

\section{Conclusion and Outlook}

DarkVesselNet provides an arXiv-ready structure for a multi-modal dark-vessel detection project. The current code validates core operators and interfaces. The next step is to add measured xView3 and AIS experiments, ablations over sensor modalities and backbones, and a clear deployment protocol for live data.

\bibliography{refs}
\end{document}